# Computational valency lexica and Homeric formularity

Barbara McGillivray (King's College London), Martina Astrid Rodda (University of Oxford)

## Abstract

Distributional semantics, the quantitative study of meaning variation and change through corpus collocations, is currently one of the most productive research areas in computational linguistics. The wider availability of big data and of reproducible algorithms for analysis has boosted its application to living languages in recent years. But can we use distributional semantics to study a language with such a limited corpus as ancient Greek? And can this approach tell us something about such vexed questions in classical studies as the language and composition of the Homeric poems?

Our paper will compare the semantic flexibility of formulae involving transitive verbs in archaic Greek epic to similar verb phrases in a non-formulaic corpus, in order to detect unique patterns of variation in formulae. To address this, we present AGVaLex, a computational valency lexicon for ancient Greek automatically extracted from the Ancient Greek Dependency Treebank. The lexicon contains quantitative corpus-driven morphological, syntactic and lexical information about verbs and their arguments, such as objects, subjects, and prepositional phrases, and has a wide range of applications for the study of the language of ancient Greek authors.

## 1. Verbal valency and valency lexicons

Valency, intended as the number and type of arguments occurring with predicates, is a central property of verbs (Tesnière 1969: 238 ff.). The English ditransitive verb *give*, for example, requires three arguments: one expressing the person giving, one expressing the object given, and one expressing the recipient. If we know that an English sentence contains an active form of the verb *give*, we can expect to find its three arguments realized in the sentence, as we can see in (1) or (2). The transitive verb *print*, on the other hand, only requires two arguments (the person or object printing, and the object being printed), so we can expect to see two arguments if a sentence contains the verb *print* in its active form (3). On the other hand, an adjunct like *yesterday* can occur with most verbs and its presence cannot be expected based only on the presence of a verb like *give* or *print* (1 and 2).

*I gave you the phone yesterday.* (1)

*He gave the receipts to the customers.* (2)

*They printed the paper yesterday.* (3)

The concept of valency is central to Dependency Grammar and is known with different terms in different linguistic subfields, including "argument structure" and "subcategorization". In this article, following McGillivray (2014: 31 ff.), we adopt an operational definition of valency, based on corpus and distributional methods, and take a theory-agnostic view on this topic. In this paper we describe AGVaLex, a corpus-driven valency lexicon for ancient Greek, and illustrate its value for historical linguistics scholarship through a case study on Homeric formulae. The lexicon was created automatically from the dependency syntax annotation of the Ancient Greek Dependency Treebank 2.0 (https://github.com/PerseusDL/treebank_data, Celano 2019) by adapting existing database queries written for Latin and described in McGillivray (2014: 31-60) and McGillivray & Vatri (2015).

Our focus on ancient languages, and particularly on ancient Greek, a "large-corpus language" (Mayrhofer,1980; Untermann, 1983), offers us the opportunity to test the effectiveness of corpus methods on a language for which no native speakers are available. This has received an increasing level of attention in recent years, in conjunction with the development and analysis of large-scale annotated corpora for corpus languages (see for example McGillivray 2014 for an overview on Latin).

The lexicon has a number of advantages and reuse potential. Thanks to its automatic creation procedure, the lexicon can be regenerated if new annotated data become available or if the annotation is corrected, which enhances its potential applications in future research. Moreover, unlike traditional dictionaries and handmade valency lexicons, computational valency lexicons like the one we present here provide a quantitative and systematic account of the valency properties of verbs reflecting the corpus they are extracted from. They can tell us whether a verb, for example, is found with a particular argument pattern, and how many times this occurs in the corpus. They can also give us information about the distribution of authors, genres and works of these patterns, and whether there is a change over time. The lexicon provides information about the number and type of arguments of all verbs occurring in the treebank. Its entries are equipped with morpho-syntactic information, namely the case of nouns and the mood of verbs, the gender and number of nouns and adjectives, and the voice of verbs. This information is highly valuable to investigate a range of linguistic questions, from the analysis of word order patterns to the study of individual verbs' constructions. The lexicon's valency patterns also display the lemmas of the arguments, which allows for lexical-semantics studies, for example to investigate the semantic fields of the subjects of objects of verbs and how they vary by author or work.

### 1.1 Applications to ancient Greek

Valency lexicons are an extremely useful tool in linguistic research on verbal complementation patterns. For ancient Greek, an important recent contribution on the topic is Keerksmaekers (2020), who analyses language change in a corpus of Greek papyri. This study, however, also showcases how much corpus pre-processing is necessary for this kind of scholarship; AGVaLex will offer an important tool for researchers wishing to conduct research on similar topics without collating their own corpus.

To illustrate this application, we propose a case study taken from Rodda (2021), on linguistic variation in archaic Greek epic poetry. The language of early Greek epic relies extensively on formulae, repeated constructions with limited syntactic and semantic flexibility; generations of researchers have investigated the precise extent of this flexibility and how it relates to issues of oral performance and language change (Rodda 2021 provides an extended bibliography; see particularly Hainsworth 1968 for an important example of this approach, and Friedrich 2019 for some criticism). Our study will show how the application of a well-developed pre-existing resource such as AGVaLex allows for new approaches to this crucial question in Homeric studies.

## 2. Previous work

Dictionaries typically display some information about verbal valency in their lexical entries. This is usually in the form of the grammatical case of arguments and the prepositions introducing the arguments themselves. For example, the dictionary entry for the verb αἱρέω 'to take' in the Brill Dictionary of Ancient Greek (Montanari 2015) reports that the verb occurs 'with acc. [...], with two acc. [...], τινά τινος [i.e., with the accusative and genitive] [...], with inf. [...], with ptc.', providing examples and translations for each construction. This information is normally followed by references and examples from texts illustrating the constructions. The number of examples shown is not proportional to the frequency of the constructions, and in many cases more uncommon constructions are given disproportionally more space in the entry. This is confirmed by the introduction to the *Thesaurus Linguae Latinae* (Bayerische Akademie der Wissenschaften, 2002),[1] for instance, and is common practice in other lexicographic resources.

Over time, dedicated valency lexicons have been created for specific languages. For example, Happ (1976) presents the only hand-made valency lexicon for Latin and was derived from a manual analysis of 800 verbal occurrences in Cicero's *Orationes*. Such resources offer high-quality information derived from a detailed manual analysis and are therefore very reliable. However, they suffer from the lack of completeness which we observed earlier, and which affects other handmade resources like traditional dictionaries.

Several large corpora of ancient Greek are available today, including the Ancient Greek Dependency Treebank (AGDT 2.0), the Diorisis Ancient Greek Corpus (Vatri and McGillivray 2018), PROIEL (Pragmatic Resources of Old Indo-European Languages, Haug and Jøndal 2008), SEMANTIA (Vierros 2018), TLG (Thesaurus Linguae Graecae), and the Perseus Digital Library (Bamman and Crane 2011). The number of syntactically annotated corpora is a subset of this list and includes PROIEL and the AGDT 2.0.

The increased availability of such large syntactically annotated corpora has made it possible to develop methods for extracting valency information automatically, thus supporting the creation of corpus-driven computational resources aiming at a systematic account of the valency behaviour of the verbs in the corpora. Typically, such resources are drawn from corpora provided with morpho-syntactic annotation, which usually follow the Dependency Grammar paradigm. As the annotation marks predicates and their arguments, it is then possible to automatically identify them and extract them in the form of a table or database. For an overview of computational valency lexicons and a discussion of valency vs subcategorization for Latin, see McGillivray (2014: 31 ff.), and for a description of such a lexicon for Latin see McGillivray et al. (2009), McGillivray and Passarotti (2012), Passarotti et al. (2016). Computational valency lexicons have several advantages over their manual counterparts. Because they directly rely on corpus data, they can easily show quantitative information such as the frequencies of each pattern for each verb, and link those back to the original corpus occurrences. They can also be easily expanded as the corpora they are based on grow, because they have been created programmatically.

---

[1] For an introduction to the methodology followed by the TLL, see https://www.thesaurus.badw.de/fileadmin/user_upload/Files/TLL/TLL_Flyer-2012_englisch.pdf. For a description of the typical structure of an entry in TLL see https://www.thesaurus.badw.de/en/hilfsmittel-fuer-benutzer/article-structure.html .

Only one corpus-based valency lexicon is currently available for ancient Greek, as far as we are aware: HoDeL 2 (The Homeric Dependency Lexicon 2; Zanchi et al. 2018 and Zanchi 2021), a project run at the University of Pavia.[2] HoDeL was automatically extracted from the syntactically annotated portion of the AGDT containing the Homeric poems. As explained in its guidelines,[3] for every verb in the Homeric poems, it includes its arguments, i.e. those dependents that are tagged as subjects, objects, object complements and predicate nominals. In the guidelines the authors point out that the lexicon does not contain referential null arguments and a number of consistency issues that affect the annotation of the treebank. Therefore, HoDeL has been manually edited to correct some annotation errors in the corpus, particularly on lemmatisation.

The online tool Myria (https://relicta.org/myria/), developed and maintained by Toon Van Hal and Alek Keersmaekers, is described as "a treebank-based vocabulary tool". It is part of the Pedalion project (Keersmaekers et al. 2019), a project based at KU Leuven which aims to improve the automatic syntactic analysis of ancient Greek. Myria can display information about ca. 6000 Greek words occurring at least 50 times in a corpus of literary texts (8th century BC to 1st century AD) from the Perseus and First One Thousand Years of Greek projects. For each of these words, "collostructures" are displayed: these are a pre-theoretical mix of collocations (e.g., the verb αἱρέω "to take" can occur "combined with an adverb", or "coordinating with another verb") and constructions (the same verb can occur "combined with an accusative noun", or "combined with a preposition in the genitive", specifically ἀντί "over, against, in exchange for"). The information in Myria is self-avowedly incomplete and being continuously updated.

## 3. The lexicon

AGVaLex was created from the Ancient Greek Dependency Treebank (AGDT 2.0; Celano 2019). AGDT 2.0 contains 557,922 tokens from the works listed in Table 1. AG VaLex is licensed under a Creative Commons Attribution-ShareAlike 3.0 United States License and is available on Figshare (McGillivray 2021).

| Author | Title |
| --- | --- |
| Aeschylus | *Agamemnon, Eumenides, Libation Bearers, Persians, Prometheus Bound, Seven Against Thebes, Supplian Women* |
| Aesop | *Aesop's Fables 1.1-1.50* |
| Athenaeus of Naucratis | *Deipnosophistae* |
| Diodorus Siculus | *Bibliotheca Historica* |
| Herodotus | *Histories* |

---

[2] https://studiumanistici.unipv.it/?pagina=p&titolo=ling-larl-hodel

[3] https://allegatifac.unipv.it/ziorufus/file/linguistica/Hodel_2_guidelines.pdf

| Hesiod | *Shield of Heracles, Theogony, Works and Days* |
|---|---|
| Homer | *Iliad, Odyssey* |
| Lysias | *Against Alcibiades* 1 and 2, *Against Pancleon, On the Murder of Eratosthenes* |
| Plato | *Euthyphro* |
| Plutarch | *Alcibiades, Lycurgus* |
| Polybius | *Histories* |
| Pseudo Apollodorus | *Bibliotheca* |
| Pseudo Homer | *Hymn to Demeter* |
| Sophocles | *Ajax, Antigone, Electra, Oedipus Tyrannus, Trachinae* |
| Thucydides | *History of the Peloponnesian War* |

Table 1. List of authors and works included in the AGDT 2.0 treebank and in AGVaLex.

The xml files of the so-called analytical layer of annotation of the treebank contain dependency-based syntactic trees. The treebank files were first converted into a tab-separated format via a Perl script, and then imported into a MySQL database; a series of MySQL query scripts then produced several database tables making up the lexicon. The scripts were adapted from the work done to create the Latin Dependency Treebank valency lexicon described in McGillivray (2014: 31-60). Specifically, we extracted all dependents of verbal forms labelled as 'SBJ' (subjects), 'OCOmp' (object complements), 'PNOM' (predicate nominals) and 'OBJ', which includes all other arguments, i.e. nouns and pronouns in the accusative, dative and genitive cases, prepositional phrases, infinitive verbs, subordinate clauses that can function as verbal objects such as accusative + infinitive constructions. The lexicon contains both dependents which are direct children and indirect children of verbal forms via preposition (AUXP), conjunction (AUXC), coordination (COORD) and apposing (APOS) notes. It is important to note that, because it was extracted from an annotated corpus, the lexicon does not include referential null arguments, i.e. those arguments that are required by a verb's valency structure, but are not lexically realised.[4]

| author | title | subdoc | verb | voice | sentence_id | frame | frame_fillers |
|---|---|---|---|---|---|---|---|
| Aeschylus | Persians | 754-756 | αἰχμάζω | active | 2901088 | active_SBJ[accusative] | active_SBJ[accusative]{ὁ} |
| Aeschylus | Persians | 713 | ἀκούω | middle | 2901049 | middle_OBJ[accusative] | middle_OBJ[accusative]{μῦθος} |
| Aeschylus | Persians | 703-706 | ἀνθίστημι | medio-passive | 2901046 | medio-passive_OBJ[dative],SBJ[nominative] | medio-passive_OBJ[dative]{σύ},SBJ[nominative]{δέος} |
| Aeschylus | Persians | 721 | ἀνύω | active | 2901073 | active_OBJ[infinitive],SBJ[nominative] | active_OBJ[infinitive]{περάω},SBJ[nominative]{στρατός} |
| Aeschylus | Persians | 726 | ἀνύω | active | 2901080 | active_OBJ[accusative] | active_OBJ[accusative]{τέλος} |
| Aeschylus | Persians | 744 | ἀνύω | active | 2901091 | active_OBJ[accusative],SBJ[nominative] | active_OBJ[accusative]{ὅδε},SBJ[nominative]{παῖς} |

Figure 1. A selection of six entries from AG VaLex.

Figure 1 displays six entries from AG VaLex. Each entry (or database record) corresponds to a verbal token occurrence in the AGDT and each column corresponds to each of eight different attributes of the token, which we can categorize into three main groups:

---

[4] An example of this can be found in the English sentence "I ate the sandwich and drank the water", where the verb "drank" requires both a subject and a direct object, but it is not lexically realised in the sentence.

1. Metadata: the columns "author", "title", "subdoc", and "sentence_id" contain, respectively, the name of the author, the title of the work, the passage where the verb token occurs and the identifier of the sentence in the treebank.
2. Verb token attributes: the columns "verb" and "voice" display the verb's lemma and voice, respectively.
3. Argument patterns: the columns "frame" and "frame_fillers" contain the valency information, as explained in more detail below.

Let us third consider the second entry in Figure 1. This entry corresponds to sentence 2901046 of the treebank, from *Persians* by Aeschylus, lines 703-706:

ἀλλ᾽ ἐπεὶ δέος παλαιὸν σοὶ φρενῶν ἀνθίσταται,
τῶν ἐμῶν λέκτρων γεραιὰ ξύννομ᾽ εὐγενὲς γύναι,
κλαυμάτων λήξασα τῶνδε καὶ γόων σαφές τί μοι
λέξον·

"Since dread long ingrained in your mind restrains you, cease, noble woman, venerable partner of my bed, from your tears and laments, speak to me with all frankness."[5]

The annotation of the first part of this sentence in the treebank is shown below:

<word id="1" cid="45088613" form="a)ll'" lemma="a)lla/1" postag="d--------" head="25" relation="AuxY" />

<word id="2" cid="45088614" form="e)pei\" lemma="e)pei/1" postag="c--------" head="25" relation="AuxC" />

<word id="3" cid="45088615" form="de/os" lemma="de/os1" postag="n-s---nn-" head="7" relation="SBJ" />

<word id="4" cid="45088616" form="palaio\n" lemma="palaio/s1" postag="a-s---nn-" head="3" relation="ATR" />

<word id="5" cid="45088617" form="soi\" lemma="su/1" postag="p-s----d-" head="7" relation="OBJ" />

<word id="6" cid="45088618" form="frenw=n" lemma="frh/n1" postag="n-p---fg-" head="3" relation="ATR" />

<word id="7" cid="45088619" form="a)nqi/statai" lemma="a)nqi/sthmi1" postag="v3spie---" head="2" relation="ADV" />

<word id="8" cid="45088620" form="," lemma="comma1" postag="u--------" head="2" relation="AuxX" />

...

</sentence>

---

[5] Aeschylus. Aeschylus, with an English translation by Herbert Weir Smyth, Ph. D. in two volumes. 1. Persians. Herbert Weir Smyth, Ph. D. Cambridge, MA. Harvard University Press. 1926.

According to the treebank annotation guidelines (Celano 2014), subjects are tagged as 'SBJ', other verb arguments are tagged as 'OBJ', predicate nominals are tagged as 'PNOM' and object complements are 'OCOMP'. All these elements can depend on a coordination node (tagged as 'COORD') and therefore take the suffix '_CO', or an apposition node (tagged as 'APOS') and then the suffix '_AP'. For a full explanation of the annotation, see Celano (2014). In the example sentence the verb form ἀνθίσταται governs the subject δέος (tagged as 'SBJ') and the direct object σοὶ (tagged as 'OBJ'), as indicated by the "head" attribute which links each of these two notes to the verbal form (tagged with the identifier "5").

The AGDT has 548,782 word tokens of which 95,841 have been tagged with the part of speech 'verb' or 'participle'; these correspond to 36,964 verb types. AGVaLex was extracted from this treebank and contains 71,887 entries, one for each of the verb tokens occurring with at least one argument in this corpus. Table 1 display some basic statistics of the lexicon.

| **Entity** | **Count** |
| --- | --- |
| Verb tokens (lexical entries) | 72,067 |
| Unique verb lemmas | 5077 |
| Unique frames | 7100 |
| Unique frames with lexical fillers | 43631 |

Table 1. Basic statistics of AGVaLex.

The treebank contains texts of 15 authors and 31 works. Table 2 shows the number of lexical entries for each author.

| **Author** | **Number of lexicon entries** |
| --- | --- |
| Aeschylus | 6170 |
| Aesop | 826 |
| Athenaeus | 5770 |
| Diodorus | 3445 |
| Herodotus | 4791 |
| Hesiod | 2171 |
| Homer | 30574 |
| Lysias | 1179 |
| Plato | 750 |
| Plutarch | 2886 |
| Polybius | 3357 |
| Pseudo-Apollodorus | 150 |
| Pseudo-Homer | 450 |
| Sophocles | 6296 |
| Thucydides | 3252 |
| TOTAL | 72067 |

Table 2. Number of AGVaLex's entries for each of the authors. Each entry corresponds to a verb token from the Ancient Greek Dependency Treebank.

| Frame | Count |
|---|---|
| active_OBJ[accusative] | 12563 |
| active_OBJ[accusative],SBJ[nominative] | 4519 |
| active_SBJ[nominative] | 4224 |
| active_OBJ[infinitive] | 2268 |
| active_OBJ[dative] | 1728 |
| medio-passive_OBJ[accusative] | 1624 |
| middle_OBJ[accusative] | 1544 |
| medio-passive_SBJ[nominative] | 1508 |
| active_PNOM[nominative] | 1239 |
| active_OBJ[genitive] | 1196 |
| active_PNOM[nominative],SBJ[nominative] | 1038 |
| medio-passive_OBJ[dative] | 831 |
| active_OBJ_CO[accusative] | 807 |
| active_OBJ[infinitive],SBJ[nominative] | 796 |
| active_OBJ[dative],SBJ[nominative] | 790 |
| medio-passive_OBJ[infinitive] | 752 |
| active_OBJ[accusative],OBJ[dative] | 745 |
| active_(εἰς)OBJ[accusative] | 682 |
| active_OBJ[dative],OBJ[accusative] | 599 |
| middle_SBJ[nominative] | 584 |

Table 3. Most frequent valency frames, with their frequency in the lexicon.

Table 3 shows the 20 most frequent frames in the lexicon. The most frequent frame is the pattern "active_OBJ[accusative]" which corresponds to constructions with accusative direct objects. Note that subjects in ancient Greek are not always expressed explicitly so this frame includes those cases in which the predicate is, for example, the first person singular and the subject is not expressed lexically.

### 3.1 Comparison with traditional lexicographical resources

A practical way to show the usefulness of the lexicon is to compare it with a commonly used scientific dictionary. We chose to compare it with the relatively recent Brill Dictionary of Ancient Greek (Montanari 2015, from here on GE, for Greek-English, the sigla provided by the editor), rather than the older LSJ (Liddell et al. 1996), as the Brill Dictionary highlights valency information more clearly, especially for high frequency verbs. So, for instance, the various constructions for τίθημι 'to set' are provided in cursive at the start of the dictionary entry, together with their most common translations in bold, before examples are provided in the body of the entry. Not all entries for verbs have an initial summary of their constructions, but even those which do not have it still highlight information about syntactic dependencies in cursive throughout the body.

In order to compare the constructions listed in the dictionary with those in the lexicon, we chose a small set of 5 transitive verbs, from the larger dataset that will be used in Section 4. These are very high frequency verbs with a reasonable variety of constructions: αἱρέω 'to take,' δίδωμι 'to give,' φέρω 'to bear,' βάλλω 'to throw,' and τίθημι 'to set;' given how common they are, they all have a summary of constructions in GE.

For each of these verbs, we noted down the dependency information that is given in GE, without taking note of diathesis (active vs. middle vs. passive), as the dictionary does not always break down meanings by diathesis unless a specific passive or middle meaning is involved. We then searched AGVaLex for all dependencies that are recorded for each verb, and noted which ones do not appear in the dictionary, as well as where they are attested. We made this choice because constructions that occur in a range of authors are arguably more likely to be recorded in a dictionary than constructions that are unique to one author, even in a partial sample like the one that forms the basis of the AGVaLex.

The results of this comparison are summarised in Table 4 below. The final column in this table contains the number of "collostructions" reported for the same verb in Myria (https://relicta.org/myria/, on which see **2** above). There is only limited overlap between the way Myria categorises collostructions and the way VaLex does (again, see section 2); therefore, the numbers are only reported for reference.

| Verb | Constructions recorded in GE | Constructions only in AG VaLex | Constructions only in AG VaLex that occur in more than one author | Constructions only in AG VaLex that occur at least 10 times | Collostructions in Myria |
|---|---|---|---|---|---|
| αἱρέω 'to take' | 5 | 31 | 1 | 1 | 10 |
| δίδωμι 'to give' | 9 | 27 | 7 | 2 | 12 |
| φέρω 'to bear' | 9 | 43 | 7 | 3 | 9 |
| βάλλω 'to throw' | 1 | 69 | 15 | 7 | 8 |

| τίθημι 'to set' | 9 | 60 | 12 | 5 | 12 |

Table 4: Comparison between GE, Myria, and AG VaLex on transitive verbs

As the table shows, while AGVaLex lists significantly more constructions than the dictionary, most of them are very rare and/or unique to one author, which makes them less relevant to a lexicographical resource that is meant to represent 'standard' Greek, with only limited reference to special usage. In addition to this, a large proportion of constructions that are unique to one author are unique to Homer, a phenomenon that sometimes has to do with Homeric syntax preserving traces of an archaic stage of development (see e.g. Hackstein 2010); for instance, as many as 48 of the 67 constructions listed for βάλλω involve prepositions that will at a later stage of the development of the Greek language be incorporated into the verb itself, creating individual compound verbs that are listed as separate dictionary entries. The issue of preverbs and their lexicalisation has been explored through computational valency lexica for Latin (McGillivray 2014, ch. 6), but no discussion of the issue in ancient Greek using similar methods exists.

That said, even for such a small sample as the one we tested, AGVaLex does sometimes bring useful additional information. For instance, we can hypothesise that a small cluster of constructions for δίδωμι 'to give' plus dative and infinitive, which only appears in Herodotus (two times), Hesiod (once), Homer (27 times) and Athenaeus (two times), is a feature of the Ionic dialect which is shared by all these authors.[6] Sometimes, the dictionary misses out on the fact that a verb can occur with a whole extra case: αἱρέω 'to take' occurs with an object in the genitive 15 times in Homer,[7] a construction that is not reported in GE. This sort of information is useful for traditional textual criticism, which often requires answering questions such as 'can this verb occur with x case?' AGVaLex offers a convenient database on which to search for answers to these questions, without having to manually check thousands of occurrences of one verb in a corpus of text.

The results above, of course, are not meant to show that the valency lexicon is superior to a published dictionary. Each lexical resource has its own purpose, but the valency lexicon does prove its worth in a test of its completeness against a common lexicographical resource, as well as its possibilities in relation to common philological aims like textual criticism, as detailed above.

In addition to the features described above, AGVaLex allows the user to retrieve summary data by construction (e.g., searching for all verbs that have the preposition ἐν 'in' plus the dative case), a type of search that cannot be performed even with an online dictionary such as GE, which has a limited user interface. On the other hand, AGVaLex is not suited to a language learner wanting to know about common constructions in an easily understandable way, and does not provide translations. The valency lexicon could also be used profitably to look for examples of specific structures for the purpose of writing a dictionary. As it also records the lexical information for nouns that enter in specific constructions with a verb, it is extremely useful for the type of semantic studies that will be described in Section 4.

---

[6] Athenaeus, who is not an Ionic author, contains many quotations from Ionic sources. This construction can coordinate with the more common dative + accusative construction, as shown in this example from Herodotus (Histories 1.54, i.e. subdoc 1.54 sentence_id 346 root_id 457705 in AG ValeX): Δελφοὶ δὲ ἀντὶ τούτων ἔδοσαν Κροίσῳ καὶ Λυδοῖσι προμαντηίην καὶ ἀτελείην καὶ προεδρίην, καὶ ἐξεῖναι τῷ βουλομένῳ αὐτῶν γίνεσθαι Δελφὸν ἐς τὸν αἰεὶ χρόνον.

[7] For instance in Homer, Il. 1.323 (subdoc 1.323 sentence_id 2274288 root_id 156142 in AG ValeX): χειρὸς ἑλόντ᾽ ἀγέμεν Βρισηΐδα καλλιπάρῃον.

A note on the Homeric Dependency Lexicon, the only other available verbal valency database for ancient Greek as described in Section 2. Since AGVaLex covers a much broader corpus than HoDeL, it makes little sense to directly compare the number of constructions retrievable by the two tools. We can, however, look at the number of Transitive Verb + Object phrases retrieved directly from Homer and Hesiod in Section 4, Table 5 below, and compare that with the data in HoDeL. For instance, for the verb αἱρέω 'to take', a script written for this purpose and running on the same AGDT treebank that HoDeL is based on retrieved 56 types, not tokens, of direct objects in the accusative; a HoDeL search of dependencies of the same verb retrieves 22 argument types tagged with the OBJ dependency and the ACC case. The discrepancy can be partially explained with the fact that the sample in Section 4 also includes the Hesiodic poems, but it still seems that the *ad hoc* script captured more occurrences than HoDeL; the outputs of the script have also been screened by hand.

## 4. Case study: semantic variation in TrV+Obj formulae

### 4.1 Aims and context of the study

The case study introduced here aims to assess the scope of semantic variation in a sample of epic formulae, and then to compare the results with a baseline corpus (for the importance of this step see Wulff, 2008). We will use Distributional Semantics to quantify semantic variation. The target of analysis is a sample of formulae made of a transitive verb and its direct object in the accusative (from here on, TrV+Obj formulae), selected exclusively on the basis of frequency. These are phrases of the type μῆνιν ἄειδε (*mênin*[acc.sing] *aeide*[pres.imp.2s]), "sing the wrath", the first two words of the *Iliad*. We will look at the semantic range of the objects of these phrases to answer questions about the semantic behaviour of these objects depending on formulaic status: do formulaic verb objects display higher or lower restrictions to their semantic range compared to non-formulaic material? Specifically, does the existence of a formulaic phrase with a specific object promote the creation of quasi-synonymous or semantically related formulae (which would increase the average similarity of the objects) or does it have a pre-emptive effect, analogous to what we observe for idioms, where the existence of an idiom with a certain meaning actually discourages the creation of synonymous idiomatic phrases (Suttle & Goldberg, 2011)?

The study of formulaic variation has been a major topic in Homeric studies at least since the 1960s (Hoekstra, 1965, 1969; Hainsworth, 1968; Postlethwaite, 1979; Friedrich, 2019). Formulae allow for a limited amount of linguistic variation, a trait which they share with idioms and other multi-word expressions in everyday language (Kiparsky, 1976). Most recently, the behaviour of formulae has been described under the linguistic framework of Construction Grammar (Goldberg, 1995): formulae are indissoluble pairs of form and function, and the restrictions to their shape are part of this system (Bozzone, 2014; Antović & Cánovas, 2016).

We use Distributional Semantics to model the range of meaning of the formulae and non-formulaic material in this case study. As a corpus-based approach, Distributional Semantics is also particularly suited to the study of dead languages such as ancient Greek, where no speaker input can be sought. In Distributional Semantics, the meaning of a word is defined as a function of its collocates in a corpus: words that share a linguistic context are also related in meaning (Harris, 1954; Fabre & Lenci, 2015). Shared linguistic contexts are modelled mathematically via word vectors which encode the frequency of co-occurrences between each word in the corpus and each of the others (with the possible exception of semantically empty "stop-words"). These vectors form a distributional space model of meaning (DSM); the distance between the vector associated to each word and the vector associated to another represents the similarity between the words' meanings.

For this case study, we will use a DSM built from the Diorisis Ancient Greek corpus (Vatri & McGillivray, 2018) using DISSECT (Dinu, Pham & Baroni, 2013). The DSM was optimised specifically for ancient Greek (Rodda, Probert & McGillivray, 2019).

## 4.2 Data and methods

The data on TrV+Obj formulae was extracted by running a Python script[8] on texts from the Ancient Greek and Latin Dependency Treebank (AGLDT: Bamman & Crane, 2011), a syntactically parsed corpus that is part of the Perseus Project. The TrV+Obj pairs were extracted from the four main archaic Greek epic texts: Homer's *Iliad* and *Odyssey* and Hesiod's *Theogony* and *Works and Days* (from here on, "the epic corpus"). Two formular editions (Pavese & Venti, 2000; Pavese & Boschetti, 2003), which are designed to mark material in the target texts as formulaic or non-formulaic based on their frequency in the texts, were used to establish which of these automatically extracted phrases are properly formulaic, i.e. repeated in the traditional language. Out of the 6764 formulaic TrV+Obj pairs that were thus extracted, only the objects of those verbs that occur at least 50 times in the epic corpus were selected, for a total of 26 verbs and 2703 tokens (ranging from 335 to 50).

The non-formulaic data for comparison was extracted from AG VaLex. All texts from the lexicon's database were included apart from those which overlap with the epic corpus, i.e. the *Iliad* and the *Odyssey*. We looked up each of the 26 target verbs in the lexicon, and manually selected the accusative objects from the existing data.

The analysis below is not on tokens, but on types (for the reasons see Barðdal, 2008), i.e. on unique object lexemes of each transitive verb. We therefore discarded any verbs that had less than 10 object types in either the epic corpus or the comparison corpus, which reduces the sample to 15 verbs. The final list of verbs, with their type frequency, is provided in Table 5; see Section 3.1 for a comparison between these numbers and the numbers that can be extracted from HoDeL.

|   | **Verb** | **Epic** | **Baseline** |
|---|---|---:|---:|
| 1 | ἔχω *ekhō* "have" | 91 | 485 |
| 2 | αἱρέω *haireō* "take" | 56 | 80 |
| 3 | δίδωμι *didōmi* "give" | 58 | 90 |
| 4 | εἶπον *eipon* "say" | 28 | 49 |

---

[8] All scripts for Section 4 are available at https://github.com/MartinaAstridRodda/dphil-thesis.

| 5 | φέρω *pherō* "carry" | 41 | 120 |
| 6 | βάλλω *ballō* "throw" | 49 | 24 |
| 7 | εἶδον *eidon* "see" | 43 | 87 |
| 8 | τίθημι *tithēmi* "set down" | 45 | 101 |
| 9 | οἶδα *oida* "know" | 29 | 53 |
| 10 | χέω *kheō* "pour" | 13 | 13 |
| 11 | ἄγω *agō* "lead" | 33 | 92 |
| 12 | λύω *lyō* "loosen" | 14 | 31 |
| 13 | τίκτω *tiktō* "give birth" | 13 | 25 |
| 14 | ἵημι *hiēmi* "send out" | 19 | 13 |
| 15 | ἀκούω *akouō* "hear" | 11 | 48 |

Table 5: Target verbs and their object types in the epic and baseline corpus, ordered by token frequency in the epic corpus (not by type frequency).

For each verb, therefore, we have a list of object types in the epic corpus and one in the baseline corpus, for a total of 30 lists. To assess their semantic similarity, we measured the cosine distance between the objects in each list and their respective centroids in the semantic space (see again Rodda, Probert & McGillivray 2019 for another example of this approach). This gives us 30 distributions of distances, which can be compared to each other or assessed for the influence of other factors.

### 4.3 Results

To assess the relationship between formulaic and non-formulaic verb phrases, we compared the semantic range of objects in the epic corpus vs. the baseline corpus for each verb. The results of this comparison are detailed in the boxplot in Figure 2.[9]

---

[9] All figures and tables in this section are reproduced from Rodda (2021).

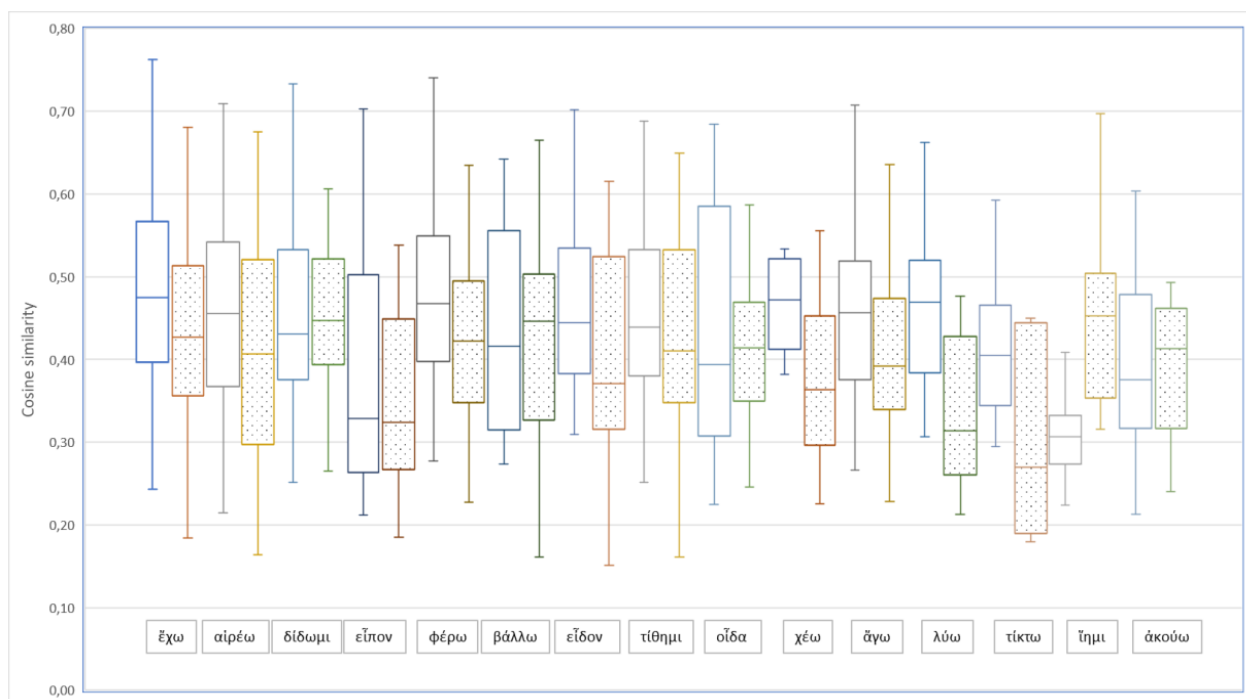

Figure 2: Box-and-whiskers plots of object similarities in non-formulaic (white) vs. formulaic (dotted) TrV+Obj pairs.

The distributions of distances for each pair were compared using the Kolmogov-Smirnov test in R (R Core Team, 2017). Two significance thresholds were set: p < 0.05 for high significance (**) and p < 0.1 for low significance (*). The results are summarised in Table 6.

| Verb | Median similarity | | Variance | | Significance |
|---|---|---|---|---|---|
| | Formulaic | Baseline | Formulaic | Baseline | |
| ἔχω | 0.427 | 0.482 | 0.0109 | 0.0100 | * (p = 0.084) |
| αἱρέω | 0.407 | 0.455 | 0.0172 | 0.0123 | (p = 0.240) |
| δίδωμι | 0.447 | 0.430 | 0.0077 | 0.0130 | (p = 0.911) |
| εἶπον | 0.324 | 0.329 | 0.0116 | 0.0188 | (p = 0.538) |
| φέρω | 0.422 | 0.468 | 0.0091 | 0.0086 | (p = 0.414) |
| βάλλω | 0.446 | 0.416 | 0.0139 | 0.0147 | (p = 0.716) |
| εἶδον | 0.371 | 0.445 | 0.0137 | 0.0103 | (p = 0.106) |
| τίθημι | 0.410 | 0.439 | 0.0133 | 0.0093 | (p = 0.723) |
| οἶδα | 0.414 | 0.394 | 0.0070 | 0.0227 | (p = 0.537) |
| χέω | 0.364 | 0.472 | 0.0096 | 0.0027 | ** (p = 0.034) |
| ἄγω | 0.392 | 0.456 | 0.0095 | 0.0099 | (p = 0.329) |
| λύω | 0.314 | 0.470 | 0.0080 | 0.0072 | (p = 0.168) |
| τίκτω | 0.270 | 0.405 | 0.0122 | 0.0062 | * (p = 0.052) |
| ἵημι | 0.453 | 0.307 | 0.0104 | 0.0027 | * (p = 0.053) |
| ἀκούω | 0.413 | 0.375 | 0.0059 | 0.0118 | (p = 0.787) |

Table 6: Comparison between formulaic and non-formulaic distributions of objects in the semantic space.

There are relatively few significant differences here, even with a higher than usual significance threshold. The four verbs that show a significant difference are ἔχω *ekhō* "have" (our most frequent verb), χέω *kheō* "pour", τίκτω tiktō "give birth" and ἵημι hiēmi "send out" (three verbs with much lower type and token frequency). For the first three, the median similarity is higher in the baseline than the formulaic corpus; the variance is always higher in the formulaic corpus.

In other words, there is only a very limited effect of formularity on semantic range, but as far as an effect can be observed, it appears to go in the direction of constructional pre-emption: objects of formulaic phrases tend to show lower semantic similarity. This is somewhat surprising, as discussions of formulaic systems (from Parry 1930 and 1932 onwards) have stressed the fact that having a range of expressions that are similar in meaning but have different metrical shapes, a result which could be easily obtained by varying lexical items and using synonyms or near-synonyms. It is possible that the definition of formularity adopted in this study, which was based on simple repetition and did not take metre into account, does not capture subtleties in the actual relation between verbs and objects which could help explain our results. It is also possible that a different approach to the data analysis would reveal a different pattern – for instance, if we set out to look for individual clusters of closely-related words among the objects of a formulaic verb, rather than measure their semantic proximity to a centroid in the semantic space. All of these avenues remain open for further analysis. What we can say for certain is that there appears to be more to be explored when it comes to the semantics of verbal constructions in early Greek epic.

## 5. Discussion and conclusions

We have presented AGValex and illustrated, via the case study in Section 4, how it can be used to explore crucial issues in ancient Greek linguistics, including issues that are primarily of interest to literary scholars, who are particularly likely to appreciate a pre-compiled dataset that can be applied in their work. The limited space devoted to the application of AGVaLex in Section 4 should not obscure the fact that the existence of the database in practice enabled this research in the first place: gathering the data on TrV + Obj constructions in Homer and Hesiod required weeks of work,[10] which would have needed to be scaled up to the entirety of the baseline corpus, a practically insurmountable task. While the results of the case study should be seen as preliminary when it comes to furthering our understanding of semantic variation in formulae, they show the promising value of the Distributional Semantics approach and of the use of a comparison database to assess how formulaic behaviour differs from non-formulaic usage.

A resource such as AGVaLex, if maintained and kept up to date, can enable research that would otherwise require more time and computational power than the average literature scholar can be expected to apply. As the availability of syntactically annotated corpora expands, these resources can be integrated into AGVaLex, ensuring the widest distribution of the data.

---

[10] New resources such as HoDeL (see Section 2 above) can speed up this task as well, but were unfortunately not yet available when the data was originally gathered. This only goes to show how AGVaLex fits in a broader trend of making Digital Humanities resources available to researchers outside of DH.